\def\year{2019}\relax
\documentclass[letterpaper]{article} 
\usepackage{aaai19}  
\usepackage{times}  
\usepackage{helvet}  
\usepackage{courier}  
\usepackage{url}  
\usepackage{graphicx}  
\usepackage{amsmath}
\usepackage{amsfonts}
\usepackage{multirow}
\title{Re-evaluating ADEM: A Deeper Look at Scoring Dialogue Responses}
\author{Ananya B. Sai\footnotemark[1]\footnotemark[2]\footnotemark[4],
Mithun Das Gupta\footnotemark[3],
Mitesh M.\ Khapra\footnotemark[1]\footnotemark[2],
Mukundhan Srinivasan\footnotemark[4]
\\ 
\footnotemark[1]Department of Computer Science and Engineering, Indian Institute of Technology, Madras\\
\footnotemark[2]Robert Bosch Center for Data Sciences and AI (RBC-DSAI), Indian Institute of Technology, Madras\\
\footnotemark[3]Microsoft, India\\
\footnotemark[4]NVIDIA, India\\
\{ananyasb,miteshk\}@cse.iitm.ac.in, migupta@microsoft.com, msrinivasan@nvidia.com}
\begin{document}
\maketitle
\begin{abstract}
Automatically evaluating the quality of dialogue responses for unstructured domains is a challenging problem. ADEM~(Lowe et al. 2017) formulated the automatic evaluation of dialogue systems as a learning problem and showed that such a model was able to predict responses which correlate significantly with human judgements, both at utterance and system level. Their system was shown to have beaten word-overlap metrics such as BLEU with large margins. We start with the question of whether an adversary can game the ADEM model. We design a battery of targeted attacks at the neural network based ADEM evaluation system and show that automatic evaluation of dialogue systems still has a long way to go. ADEM can get confused with a variation as simple as reversing the word order in the text! We report experiments on several such adversarial scenarios that draw out counterintuitive scores on the dialogue responses. We take a systematic look at the scoring function proposed by ADEM and connect it to linear system theory to predict the shortcomings evident in the system. We also devise an attack that can fool such a system to rate a response generation system as favorable. Finally, we allude to future research directions of using the adversarial attacks to design a truly automated dialogue evaluation system.
\end{abstract}

\section{Introduction}
AI agents capable of having human-like conversations find various applications such as providing an automated help-desk for customer service and technical support, serving as language learning tools, personal assistants and as a source of entertainment/recreation. The research community has accessibility to a number of datasets for the task of dialogue generation \cite{DBLP:conf/naacl/RitterCD10,DBLP:conf/sigdial/LowePSP15,DBLP:conf/aaai/SahaKS18}. This has led to the emergence of goal-driven as well as non-goal-driven conversation models \cite{DBLP:conf/emnlp/RitterCD11,DBLP:journals/corr/VinyalsL15,DBLP:conf/emnlp/WenGMSVY15,DBLP:conf/naacl/SordoniGABJMNGD15,yu2017seqgan,li2017adversarial}. However it is hard to measure the scientific progress towards a conversational agent due to the lack of good evaluation metrics. 
Human evaluations and comparisons are reported in most of the works on samples of the data. However, it is infeasible to have a human in the loop to give feedback and scores for training and evaluating dialogue systems. This has led to adoption of the existing automatic evaluation metrics for the task of scoring the generated dialogues. Popular word overlap based metrics such as BLEU \cite{DBLP:conf/acl/PapineniRWZ02}, METEOR \cite{DBLP:conf/acl/BanerjeeL05}, ROUGE \cite{lin2004rouge} and the various word embedding based metrics have been used to score dialogue generation systems. However, they can't handle the diversity of the range of valid responses and are shown to correlate poorly with human judgement \cite{DBLP:conf/emnlp/LiuLSNCP16}. 

Consider the conversation in Table \ref{tab:conv}. All of the responses sound reasonable and logical given the context. However, one can notice that there is no word overlap across any pair of the responses. Should any one of the responses be considered the ground truth, a model that produces any other response will receive very low scores. The popular BLEU metric would assign the responses a score of $0$. Even the word embeddings based metrics face a similar issue. The task of dialogue generation is tough to evaluate due to the huge number of valid responses possible.

Simultaneously research efforts are underway in the direction of using generative adversarial network (GAN) based dialogue systems where the generator and discriminator are trained in parallel. The discriminator's role is to determine whether the  given dialogue is machine-generated or human-generated. However, it is tricky to use the discriminator trained this way as one cannot be sure if it is infact a good discriminator or seems good due to a poorer generator. Such training might also lead to a very model specific evaluator rather than a generic one.

\begin{table}
	\centering
	\begin{tabular}{|l  l|}
		\hline
		\textbf{Context:} & Do you want to watch a movie today?  \\
		\hline
		\textbf{Valid}& Sure, Avengers infinity war is out.\\
		\textbf{responses:} &I didn't finish my assignment.\\
		&Yeah, but will it rain tonight?\\ 
		\hline
	\end{tabular}
	\caption{Diversity of valid responses in a Dialogue}
	\label{tab:conv}
\end{table}
In this work we look at one such recently proposed metric, viz. Automatic Dialogue Evaluation Model (ADEM) by Lowe et al. (\citeyear{DBLP:conf/acl/LoweNSABP17}) which trains a neural net based model to score the overall quality of dialogue responses on a scale of 1-5. ADEM is trained on Twitter corpus with the help of AMT (Amazon Mechanical Turk) for obtaining human scores for the dialogues. ADEM uses a hierarchical RNN encoder architecture to obtain the context embedding as well as the embeddings of the model response (\textit{i.e.}, the response to be evaluated) and the reference response. The score of the response is a function of these embeddings.
The primary flaw with this is the multiplicative setting of embedding interactions which has been shown to result in embedding vectors of high conicity (lesser spread) by Chandrahas et al. (\citeyear{Chandrahas2018}). We provide an analysis of this phenomenon and verify the claim experimentally as well. 

We do a deeper analysis and show that the scores assigned by ADEM to various dialog responses are banded in a small margin around the mean of about 2.75 with a spread of around 0.34. Further analysis shows that due to the inherent flaw in its design, ADEM is actually susceptible to adversarial attacks wherein it fails to provide appropriate scores for multiple responses. 
This is counterintuitive to the claims made by ADEM, which tries to find diversified scores for varied responses correlated to human scores.
Lastly, we look at the cost function of ADEM and prove that the spread cannot be widened by the multiplicative model leading to a high conicity. 
ADEM was proposed with the motivation of addressing the limitations of metrics such as BLEU by providing high scores to diverse responses correlating human scores, but in the process it scores most of the responses extremely closely and shows very low discriminative power to handle a wide array of real life responses.

\section{Related Work}
Since our work focuses on a critique of automatic evaluation metrics we first do a quick review of various popular metrics used for automatic evaluation and then review works which are similar in idea to ours and themselves do a critique of these evaluation metrics. The research on dialogue generation models is guided by the dialogue evaluation metrics which provide the means for comparison. BLEU and METEOR scores, originally used for machine translation, are adopted for this task by various works \cite{yu2017seqgan,DBLP:conf/emnlp/RitterCD11,DBLP:conf/naacl/SordoniGABJMNGD15,DBLP:conf/emnlp/WenGMSVY15,DBLP:conf/acl/GalleyBSJAQMGD15,N16-1014}. BLEU analyses the co-occurrences of n-grams whereas METEOR creates an explicit alignment using exact matching, followed by WordNet synonyms, stemmed tokens, and paraphrases, in that order. Similarly the ROUGE metric variants, originally used for automatic summarization, work on overlapping units such as n-grams, word sub-sequences and word pairs. The ROUGE metrics being recall-oriented, require a sufficient number of references to produce reliable scores.\\

While the above metrics directly use words for comparison, another alternative is to use word embeddings. Word embeddings are calculated by methods such as Word2Vec \cite{DBLP:conf/nips/MikolovSCCD13}, GloVe \cite{DBLP:conf/emnlp/PenningtonSM14}, which represent words as vectors derived from the contexts they appear in. Using these word vectors, the \textit{Greedy Matching metric} considers each token in actual response and greedily matches it with each token in predicted response based on cosine similarity of word embedding (and vice-versa). The total score is then averaged across all words, making this greedy approach favor responses with keywords that are semantically similar to those in the ground truth response. Certain metrics compute sentence embeddings using the word embeddings. The \textit{Embedding Average metric} calculates sentence-level embeddings simply by averaging the word embeddings for each token/word in the sentence. Another way to calculate sentence-level embeddings is by using vector extrema, where for each dimension in the word vector, the most extreme value amongst all word vectors in the sentence is used in the sentence-level embedding. The basic idea is that by taking the maximum along each dimension, we can ignore the common words (which will be pulled towards the origin) and prioritize informative words which will lie further away in the vector space. Both these methods compare the ground truth response and the retrieved response by computing the cosine similarity between their respective sentence level embeddings.

These metrics face a lot of criticism  for use in NLG tasks \cite{DBLP:journals/corr/abs-1808-10192,DBLP:conf/eacl/Callison-BurchOK06,DBLP:conf/emnlp/Callison-Burch09}.
Another major shortcoming of these evaluation metrics was studied by \citeauthor{DBLP:conf/emnlp/LiuLSNCP16} (\citeyear{DBLP:conf/emnlp/LiuLSNCP16}), who show that the scores generated by these metrics correlate poorly with human judgement. They perform this analysis using Twitter dataset consisting of casual chit-chat and Ubuntu corpus containing technical conversations and conclude that metrics which do not specifically correlate with human judgements on a new task should not be used to evaluate that task. More recently, \citeauthor{DBLP:journals/corr/abs-1809-08267} (\citeyear{DBLP:journals/corr/abs-1809-08267}) add to this debate suggesting the comparison of correlation to be made at the corpus level rather than sentence level.\\
\begin{table}
	\centering
	\begin{tabular}{|c l c|}
		\hline
		\textbf{} & \textbf{Positive / Negative ex}  &\textbf{Ideal accuracy}  \\
		\hline
		1 & human / human response  & 50\%\\
		2 & machine / machine response  & 50\%\\
		3 & human / random response  & 100\%\\
		4 & human / skipped response  & 100\%\\
		\hline
	\end{tabular}
	\caption{Simple scenarios for computing ERE (Evaluator Reliability Error) \cite{li2017adversarial}}
	\label{tab:eva}
\end{table}

\citeauthor{li2017adversarial} (\citeyear{li2017adversarial}) discuss various evaluator models and architectures for dialogue evaluation including SVM, concatenation based neural network, and a hierarchical neural network. The task of the dialogue evaluator in these cases is to classify a response as human or machine-generated. Additionally, to test the reliability of an evaluator, they introduce four scenarios as shown in Table. \ref{tab:eva}, where one can know in advance how a perfect evaluator would behave. They define a score called Evaluator Reliability Error (ERE), as the average deviation of the evaluator model from the gold standard accuracies in each of these scenarios, with equal weightage to each of the four cases. However, they evaluate their dialogue generator on the hierarchical model alone, on which they report ERE as 0.193, with their generator being able to fool the evaluator 9.8\% of the times. 

In this work we evaluate ADEM which scores dialogue responses and design a battery of experiments, which not only bring out the shortcomings in ADEM, but also pave the way for re-designing such a system. We perform both theoretical and empirical evaluation of ADEM and conclude that the response evaluation systems need to address the linear system constraints as well as the semantic constraints to be uniformly useful across different dialogue domains. To the best of our knowledge, this kind of analysis of dialogue scoring systems has not been done before. 

\section{ADEM: Background}
The automatic scoring method, ADEM, proposed by \citeauthor{DBLP:conf/acl/LoweNSABP17} (\citeyear{DBLP:conf/acl/LoweNSABP17}) computes the score for a dialogue response by using a dot-product between the vector representations of the dialogue context $c$, reference response $r$, and model response $\hat{r}$
\begin{align}\label{Eq:ADEM_c1}
	score(c,r,\hat{r})=(c^TM\hat{r}+r^TN\hat{r}-\alpha)/\beta
\end{align}
where $M,N \in \mathbb{R}^n$ are learned matrices initialized to the identity and $\alpha, \beta$ are scalar constants used to initialize the model's predictions in the range [1,5].

The model is trained to minimize the squared error between the model predictions and the human score with L2-regularization.
\begin{align}
	\mathcal{L}=\sum_{i=1:K}[score(c_i,r_i,\hat{r}_i)-human_i]^2+\gamma||\theta||_2
\end{align}

Using the above formulation, the authors report a high correlation between the ADEM scores and human scores at utterance level and system level. However a closer analysis of the model uncovers multiple adversarial scenarios where the evaluator is easily fooled or does the opposite of what is expected. We show that the neural network based ADEM model does not perform well with respect to the syntax and semantics of the response to be scored. The model is not able to disambiguate 
whether the words in a response are jumbled up or if random words of the response have been repeated. We explore the possibility of adversarial attacks and fooling mechanisms on dialogue evaluator models and propose a 
systematic approach 
to target dialogue evaluation systems. Using our method, we find irrelevant responses for the given context which ADEM scores high, concluding 
that ADEM is susceptible to adversarial attacks leading to significant loss in its scoring ability.\\ 

Our experiments show a consistent tendency in ADEM to score all dialogues very close to a mean value of $2.75$ on a $5$-point scale with $0.34$ standard deviation.
Further, we compute the Conicity (defined as the mean of the cosine similarities of each of the vectors with the mean vector \cite{Chandrahas2018}) of the response embeddings computed by ADEM. On a dataset of 200k such responses, the conicity value of the response embeddings generated by ADEM is $\approx 0.6$ indicating a very low vector spread in the embedding space. Exploring further for explanations for this, we look into ADEM's score formulation and highlight our observations. Through our analysis, we intend to provide guidelines for evolution of future dialogue evaluation metrics.

\subsection{Deeper Look into ADEM's Scoring Function and the Dialogue Embeddings}
\begin{table}
	\centering
	\resizebox{\columnwidth}{!}{
	\begin{tabular}{|l c c|}
		\hline
		\textbf{Response to be evaluated} & \textbf{ADEM mean} & \textbf{ideal score} \\
		\hline
		ground-truth response & 2.75 & 5 \\
        context repeated as response & 3.03 & 1 \\
		machine generated response$^*$ & 2.64 & 1\\
		swapping reference response & 2.6 & 5 \\
         and machine response & & \\
		\hline
	\end{tabular}
	}
	\caption{ADEM scores on simple test cases ($^*$Machine generated responses are obtained by training a GAN based neural dialogue generation model \cite{li2017adversarial})}
	\label{tab:scores1}
\end{table}
Borrowing the idea of the ERE metric proposed by \citeauthor{li2017adversarial} (\citeyear{li2017adversarial}), we evaluate ADEM on certain similar straightforward cases with easy-to-guess desired outcomes. While the ERE metric deals with a binary classification of dialogues as human versus machine-generated, we adopt the idea for a score based evaluators like ADEM. This analysis is presented in Table \ref{tab:scores1} using the Microsoft Research Social Media Conversation Corpus by \citeauthor{DBLP:conf/naacl/SordoniGABJMNGD15} (\citeyear{DBLP:conf/naacl/SordoniGABJMNGD15}). We first check ADEM's evaluations of the reference responses. One would expect the scores on good reference response to be high. Note that the reference responses for dialogues in this dataset have high human ratings.
Similarly, when the context itself is repeated back as the response, it forms an unnatural reply in most cases and one can expect scores to be very low. Next we check ADEM's scores in the case of the machine generated responses. We train a GAN based neural dialogue generation model based on \citeauthor{li2017adversarial} (\citeyear{li2017adversarial}) for a very limited number of $5$ epochs before using the generated responses for evaluation. We deem this training to be insufficient and hence expect an ideal evaluator's scores to be low. Finally we think of the case where the reference response and the responses-to-be-evaluated (in this case, the machine generated responses) are swapped. The idea is to find out how much the reference response influences the evaluation of a response. In the numerous dialogue datasets which often involve some automation in their creation, it would not be uncommon to find certain mediocre reference responses. A human evaluator merely uses the reference response as an unabiding guideline and can recognize a better-suited response without being misled/distracted by a bad reference response. We hence suggest an ideal score of 5 for this scenario. In all of these cases, we find ADEM's score to be quite similar. We find that ADEM has a mean score between $2.6-3$ in all of these cases with a very low standard deviation ranging between $0.32-0.36$. We also find $60-71\%$ of the scores within 1 standard deviation of the mean.

We perform analysis to explain the possible reasons for the clustering of the scores around the mean as produced by ADEM. To proceed further with our analysis we borrow some of the concepts defined by Chandrahas et al. (\citeyear{Chandrahas2018}).
\begin{figure}[ht!]
\includegraphics[width=.95\columnwidth]{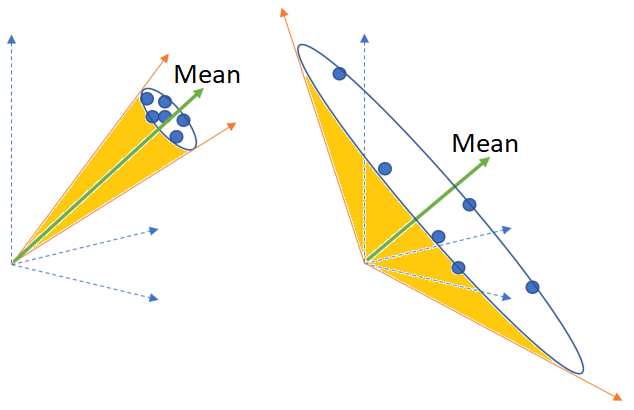}
\caption{Conicity for a set of vectors. Left: high conicity with a small vector spread obtained from multiplicative systems. Right: low conicity with a large vector spread obtained from additive systems.}
\label{Fig:conicity}
\end{figure}
They define two broad classes of models which encode the behavior of embedding vectors, namely:
\begin{itemize}
\item multiplicative systems
\item additive systems
\end{itemize}
For a triplet of embedding vectors, (the context, reference and response in our case,) multiplicative systems take the form 
\begin{equation}
\sigma_m (c, r, \hat{r}) = \hat{r}^T f(c,r)
\end{equation}
whereas an additive system takes the form 
\begin{equation}
\sigma_a (c, r, \hat{r}) = \hat{r} +  f(c,r)
\end{equation}
Note that the function $f$ can be selected such that the system is uniformly multiplicative or additive for all the components, or vice-versa, wherein two of the components interact additive and the remaining have multiplicative interactions. For ADEM this function $f$ can be written as
\begin{equation}
f(c,r) = A^T c + B^T r
\end{equation}
which is a linear additive function. Given a set of embedding vectors $V = [v_1, v_2, \ldots]$, the two metrics from Chandrahas et al. (\citeyear{Chandrahas2018}), are defined as
\begin{itemize}
\item Alignment to Mean (ATM)
\begin{align*}
	\mathrm{ATM}(v, \bar{V})= v^T \bar{V}
\end{align*}
where $\bar{V}$ is the mean of all the vectors in $V$.
\item Conicity of the set of vectors is the mean of the ATMs for all the vectors in the set
\begin{align*}
	\mathrm{Conicity}(V)= \frac{1}{|V|}\sum_{x \in V} \mathrm{ATM}(x^T \bar{V})
\end{align*}
\end{itemize}
Chandrahas et al. (\citeyear{Chandrahas2018}) empirically show that multiplicative systems lead to embedding spaces with high conicity and hence all the responses get closely huddled around the mean as shown in Fig.~\ref{Fig:conicity}.

The principal finding from our analysis so far points towards the fact that the scores generated by ADEM are all tightly clustered around the mean value and hence the discriminative power of the system is limited at best. Taking a deeper look at the ADEM cost function we can write
\begin{eqnarray}
  score(c,r,\hat{r}) &=&(c^TM\hat{r}+r^TN\hat{r}-\alpha)/\beta \\  
                     &=& a^T\hat{r} - b
\end{eqnarray}
where $a = (c^TM + r^TN)/\beta$ and $b = \alpha/\beta$. Without loss of generality, assume that there are two model responses $\hat{r}_1$ and $\hat{r}_2$ for the same context $c$ and reference $r$. Since the model is linear and all other variables are fixed, for a new (valid) response $\hat{r}_n$ which lies on $\hat{r}_2-\hat{r}_1$ (or more generically a linear combination of $\hat{r}_1$ and $\hat{r}_2$), we can write the new score as

\begin{eqnarray}
  s_n &=& a^T\hat{r}_n -b\\  
      &=& a^T \left( \hat{r}_1 + \Delta \frac{\hat{r}_2 - \hat{r}_1}{M} \right) -b \\
      &=&  \left( s_1 + \frac{\Delta}{M} (s_2 - s_1) \right) - b 
\end{eqnarray}
where $\Delta$ is a small step in the direction of $\hat{r}_2-\hat{r}_1$, $M$ is the magnitude of the vector $\|\hat{r}_2-\hat{r}_1\|$. Assuming $s_2 > s_1$, the score difference $(s_n - s_1)$ can be maximized (leading to diverse scores for close vectors), by taking $M\rightarrow 0$ which leads to $\hat{r}_1 \rightarrow \hat{r}_2$ which leads to mapping all the response vectors very close to each other leading to high conicity and low vector spread. 

Our analysis for ADEM points to a similar artifact.
ADEM's response embeddings of $200k$ twitter responses has a conicity value of $0.59$. Further we note that the score is a linear formulation [see Eq.~\ref{Eq:ADEM_c1}]. This means that for a given context and reference response, there is no way to score two valid responses high by circumventing any bad embeddings that lie in the subspace spanned by these two responses. 

\section{Analysing Robustness of ADEM to Adversarial Attacks}
\begin{table}
	\resizebox{\columnwidth}{!}{
	\begin{tabular}{|l c c c|}
		\hline
		\textbf{Response to be evaluated} & \textbf{mean} &\textbf{SD} & \textbf{\%1 SD}\\
		\hline
		Reference response & 2.75 & 0.34 & 71.65\\
		Punctuation removed & 2.85 & 0.31 & 71.65\\
		NLTK stopwords removed & 2.69 & 0.33 & 70.60\\
		25 common stopwords removed & 2.80 & 0.24 & 69.08\\
		{[pro]}nouns and verbs only & 2.80 & 0.36 & 68.96\\
		Named entities removed & 2.74 & 0.35 & 70.60\\
		Replace words with synonyms & 2.83 & 0.32 & 70.36\\
		Jumble words in the sentence & 2.73 & 0.33 & 72\\
		Reverse the response & 2.75 & 0.33 & 68.84\\
		Retain only nouns & 2.73 & 0.39 & 68.26\\
		Repeat words in the response & 2.70 & 0.36 & 71.41\\
		Generic and Irrelevant responses:& & & \\
		I'm sorry, can you repeat? & 2.65 & 0.34 & 69.43\\
		I will do & 2.69 & 0.34 & 70\\
		fantastic! how are you? & 3.18 & 0.4 & 69.4\\
		\hline
	\end{tabular}
	}
	\caption{ADEM scores on simple dataset variants. The last column indicates the percentage of scores within one standard deviation of the mean score.}
	\label{tab:scores}
\end{table}
For the following section of human constructed attacks, we run the experiments on the Microsoft Research Social Media Conversation Corpus \cite{DBLP:conf/naacl/SordoniGABJMNGD15}. This corpus contains a curated list of 3 turn twitter conversations, all of which received an average score of 4 or higher for good response quality by 3 crowd-sourced annotators apiece. Human scoring of the dialogues was performed on a 5-point scale similar to ADEM's automated scoring on a scale {[1-5]} which makes it easier for comparison.\\
We construct some simple test cases by modifying the reference response and summarize the findings in Table \ref{tab:scores}. This analysis on ADEM shows:
\begin{itemize}
\item multiple adversarial scenarios where the evaluator is easily fooled or does the opposite of what is expected
\item the ADEM score in most cases is a conservative average value of around 2.75
\end{itemize}
Although these experiments are reported on ADEM model, they can be used to analyze any other dialog evaluator model. Table \ref{tab:scores} and Table \ref{tab:corr} show the results of the experiments with various dataset modifications of different flavors explained below:\\
\begin{itemize}
\item \textbf{Evaluation of the Ground Truth Response} to estimate the reliability of the evaluator. We evaluate the scores on the responses in the training set of MSR Social Media Conversation corpus, which are already rated high by human annotators. This first experiment revealed that the ADEM scores are concentrated around the average value of 2.75 which seems like a conservative middle value on a scale of {[1-5]}. The standard deviation of the scores is 0.34 with 71.65\% of the scores within the first standard deviation.
\item \textbf{Slight Modifications} such as removing punctuation and stopwords should probably lower the score only a little for a given response.
\subsubsection{Removing Punctuation}
In this modification of the dataset, the test responses are stripped of all punctuation marks. We note that a human's score of such a response would still be just about the same. We conclude this from an human evaluation of a sample of $150$ such responses by crowd-sourced annotators resulting in an average score of $4.83$. However, the correlation of the scores by ADEM on responses with and without this modification is not that strong.
\subsubsection{Removing Stopwords}
Removing stop words should not affect the score much if one is lenient with the grammar. The standard English stop words from NLTK (Natural Language ToolKit) library of python were removed in this variant (NLTK stopwords removed). Another variant where only the top 25 most common stop words were removed in also listed in Table \ref{tab:scores} as 25 common stopwords removed. The list of the 25 is as follows:
\begin{center}
	a, an, and, are, as, at, be, by, for, from,\\
	has, he, in, is, it, its, of, on, that, the,\\
	to, was, were, will, with
\end{center}
A peculiarity that can be observed is the better correlation of scores when more stopwords are removed than when just the 25 most common ones are filtered out. While this could be explained as more stopwords eliminated leading to reduction in noise and unnecessary dimensions, it is questionable whether human scoring would follow the same pattern. The human evaluation on $150$ samples of each of these types decrements average scores to $4.6$ and $4.2$ respectively when only 25 stopwords are removed and all the NLTK stopwords are removed.
However, Table \ref{tab:corr} shows that the score is better than that of the original response in over 60\% cases when either punctuation or the 25 most common stopwords are removed.
\begin{table}
	\centering
    \resizebox{\columnwidth}{!}{
	\begin{tabular}{|l c c c|}
		\hline
		\textbf{Variant} & \textbf{Pearson} & \textbf{Spearman}  &\textbf{Better score} \\
		\hline
		Punctuation removed & 0.55 & 0.5 & 64.41\%\\
		NLTK stopwords removed & 0.78 & 0.76 & 37.92\% \\
		25 common stopwords removed & 0.6 & 0.57& 60.33\%\\
		{[pro]}nouns and verbs & 0.52 & 0.49 & 56.71\% \\
		Named entities removed & 0.98 & 0.97 & 11.2\% \\
		Replace words with synonyms & 0.79 & 0.75 & 68.03\%\\
		Jumble words in the sentence & 0.68 & 0.64 & 47.02\% \\
		Reverse the response & 0.52 & 0.49 & 48.66\% \\
 		Retain only nouns & 0.29 & 0.26 & 50.64\% \\
		Repeat words in the response & 0.91 & 0.90 & 37.57\%\\
        "fantastic! how are you?" & 0.34 & 0.32 & 86.93\%\\
		\hline
	\end{tabular}
	}
	\caption{Correlation of ADEM scores on different variants of the response with the ADEM scores on original (reference) response. p-values in all these cases are $<0.001$. The last column indicates the percentage of times the concerned variant received a better score than original}
	\label{tab:corr}
\end{table}
\item \textbf{Simplifying the Response:} 
\subsubsection{Retaining Only the Nouns, Pronouns and Verbs in the Response}
Our intention was to check if a simpler version of the response can be created by removing adjectives, adverbs, etc and retaining just the nouns, pronouns and verbs. We performed a human evaluation of $150$ dialogue samples to check if the core idea of the modified responses would still be the same in most cases. We find the human scores drop to an average of $2.78$. We hence observe that the hypothesis is not true for all responses and do not pose expectations on correlation of scores for this variant.
\subsubsection{Removing Named Entities} An even milder modification is to just remove the named entities which we identify using the POS (Part of Speech) tagger functions and Named Entity Recognition modules of the NLTK library. The score isn't expected to be affected much by such a modification. This is also reflected by the strong correlation in the scores of this variant with the original response.
\item \textbf{Replacing Words by Synonyms:} This variant contains most of the words replaced by their synonyms (excluding stop words and named entities) using WordNet from NLTK. Since the syntax, semantics and the meaning of the response are unaltered, the score should ideally be [almost] the same and correlation should be high. ADEM scores exhibit a decent correlation between the response and its variant with synonyms-replaced.
\item \textbf{Perturbing the Response} by reversing the sentence or jumbling the words, randomly repeating words in the response create unnatural responses.
\subsubsection{Jumble the Sentence}
If the response words appear in jumbled order, low scores are expected. Also it would be an interesting insight into the working of an evaluator model to check whether the scores on this variant correlate with the original response. From the correlation values in Table \ref{tab:corr} we observe that ADEM does not take into account the syntax and semantics of a response while scoring it.
\subsubsection{Reverse the Sentence}
This can be considered as a special case of the previous variant. The word order is the exact reverse of the original.
\subsubsection{Retain Only Nouns in the Response}
Using the POS tagger functionality of NLTK library, the utterance was modified to contain only the nouns (proper and common nouns). This should ideally obtain low scores from a dialogue evaluator. This variant indeed elicits the weakest correlation with the original response which seems logical as the meaning of the response is lost.
\subsubsection{Repeat Words in the Response}
An unnatural variant of a response/utterance is created in this case by randomly choosing $m/2$ words to be repeated in the utterance, where $m$ is the length of the utterance. An evaluator should be able to distinguish when a repetition is supposed to be penalized and when it is natural. Whereas, it can be observed that the scores on the response with repetitions are strongly correlated to the original responses.
\item \textbf{Generic and Irrelevant Responses}
Table \ref{tab:scores} also shows the scores when a generic response is used, such as "I'm sorry, can you repeat?" which would be applicable as a suitable response in most cases. Also a dialogue evaluation metric should not possess a loophole where a dialogue generation system can learn to pick up a certain utterance in the vector space with which it can consistently manage to get high scores from the evaluator. The response "Fantastic! How are you?" seems to be one such case with ADEM, where it is given relatively higher scores in any context. It was found to be scored higher than the reference response by ADEM in $86.93$\% of the cases.
\end{itemize}

Although these are very simple cases, the goal towards a good dialogue evaluation model would be realized only with the ability to handle such cases.

\section{Whitebox Attack on ADEM}
While a lot of work has been done around fooling convolutional networks using gradient descent, 
the methods of adding noise to fool an NLP model is not straightforward due to the discrete nature of text. Works towards this goal hence try various ways of perturbing the data or producing adversarial/unfavorable examples. Targeting the Question Answering models, \citeauthor{DBLP:conf/emnlp/JiaL17} (\citeyear{DBLP:conf/emnlp/JiaL17}) insert adversarial sentences in the SQuAD dataset \cite{DBLP:conf/emnlp/RajpurkarZLL16}, which do not affect the correct answer or mislead humans. However, they report a drop in the average accuracy of sixteen published models from 75\% to 36\% F1 score. Li et al. (\citeyear{DBLP:journals/corr/LiMJ16a}) propose to interpret the importance of various aspects of a neural network by analyzing the effects of erasing various parts of the representations such as input word-vector dimensions, intermediate hidden units and input words. \citeauthor{DBLP:conf/ijcai/0002LSBLS18} (\citeyear{DBLP:conf/ijcai/0002LSBLS18}) fool text classification by either adding, removing or modifying the inputs. Zhao et al. (\citeyear{zhao2018generating}) add perturbations in the continuous space rather than at the discrete input level, using autoencoders to map the discrete text into continuous codes.\\

\begin{table}
	\centering
	\begin{tabular}{|l c c c c|}
		\hline
		\textbf{Type} & \textbf{mean} & \textbf{std dev}  &\textbf{max} &\textbf{min} \\
		\hline
		Original responses & 2.75 & 0.34 & 4.19 & 1.44\\
		Brute force search & 3.93 & 0.24 & 4.7 & 3.39\\
        Annoy index search & 3.62 & 0.29 & 4.3 & 3.24\\
		\hline
	\end{tabular}
	\caption{Statistics of the ADEM scores on various responses}
	\label{tab:brutenannoy}
\end{table}
\begin{table}
	\centering
    \scriptsize
    \resizebox{\columnwidth}{!}{
	\begin{tabular}{|p{.33\columnwidth} | p{2cm} c c|}
		\hline
		\textbf{Context} & \textbf{Response} & \textbf{ADEM}  &\textbf{human} \\
		\hline
		\multirow{5}{.33\columnwidth}{$<$first\_speaker$>$ Hi,\\how are you doing?\\$<$second\_speaker$>$ \\not too bad, getting\\ over a cold} & 
        colds are no good ,hope it clears away soon. & 2.9 & 5\\
        \cline{2-4}
		& $<$first\_speaker$>$ sure, you can get an autographed one right here: $<$URL$>$ & 3.9 & 1\\
        \cline{2-4}
        & $<$first\_speaker$>$ mate, get the app, you can watch while running, & 3.79 & 2.5\\
		\hline
        \multirow{5}{.33\columnwidth}{$<$first\_speaker$>$Awesome\\! Did you graduate high\\school this year ?\\$<$second\_speaker$>$ Nope\\2 more years !} & 
        $<$first\_speaker$>$ Wow ! Going to be a junior . You youngster ! & 2.4 & 5\\
        \cline{2-4}
		& $<$first\_speaker$>$ that is what sundays are for ! & 3.6 & 1\\
        \cline{2-4}
        & $<$first\_speaker$>$ better safe than sorry ! & 3.56 & 1\\
        \hline
	\end{tabular}
	}
	\caption{Examples with ADEM and human scores}
	\label{tab:brute_scores}
\end{table}
{
\centering
\begin{table*}
\centering
	\begin{tabular}{|l l l l|}
		\hline
		\textbf{Context} & \textbf{Reference response} & \textbf{Response to evaluate} & \textbf{ADEM}\\
		\hline
		do you want to watch &  will it rain tonight ? &  will it rain tonight ? & 2.22 \\
         a movie today ?& & yes but will it rain tonight ?& 2.52\\
         & & sure , avengers infinity war is out & 2.37\\
		& & i did not finish my assignment & 2.49\\
		&  &  tonight rain it will ? & 2.37\\
		& & ? tonight rain it will & \textbf{2.7}\\
		\hline
	\end{tabular}
	\caption{ADEM scores on an example context with various responses}
	\label{tab:q2}
\end{table*}
}
In the context of neural network based dialogue evaluators, we propose a white box attack to game the model for high scores (or low scores) given a context and a reference response. We employ guided backpropagation \cite{DBLP:journals/corr/SpringenbergDBR14} from the score function back to the response sentence embedding level of ADEM model. This directs changes in the response embedding towards producing the desired score. Employing this method, we arrive at an embedding that produces a score between $4.6-4.9$. We now need the sentence/response that translates to this computed embedding that can fetch the desired score. Here, we settle for an approximation of the desired score by finding the sentence closest to the desired sentence embedding computed. For this, we use a database of sentence embeddings constructed on 470k Twitter responses crawled using the Twitter IDs given by \citeauthor{DSTC6_End-to-End_Conversation_Modeling} (\citeyear{DSTC6_End-to-End_Conversation_Modeling}) for customer service conversations. The embeddings are indexed using Annoy index \cite{DBLP:journals/corr/LiZSWZL16}, which is designed to identify approximate nearest neighbors in a multidimensional space. We query the corpus using this index with our response embedding to obtain its [approximate] nearest neighbor embeddings. We gather 400 of these nearest neighbors and find the best score on these by ADEM. The scores increase on average by $0.87$ to form an average score of $3.62$.
To understand the upper limit or the potential for improvement in the score when using this database, we also employ brute force search on the entire embeddings database to find the best score given by ADEM. In other words, we compute the scores on all the response embeddings in our database and pick the highest. The mean score in this case is $3.93$ with an average increment of $1.18$. These results are tabulated in Table \ref{tab:brutenannoy}

\subsection{Human Evaluation}
We check the relevance of the responses that were scored high by ADEM for the given context. The human evaluation is made by in-house annotators using 2 approaches:\\
\begin{itemize}
\item  scoring the top response, as selected by ADEM, on a scale [1-5] similar to ADEM's scoring scheme
\item re-ranking ADEM's top 5 scored responses. The dialogues are presented in the order of ADEM scores (high-low). The annotators assign their rank/preference of each response for the given the context.
\end{itemize}
We find that the average human rating on the responses scored highest by ADEM is $1.9$ for $250$ samples. The re-ranking experiment could not be followed as intended as most of the responses were bad leading to no preferences in the responses by the annotators in most cases.\\
Table \ref{tab:brute_scores} shows some examples of human and ADEM scores on original responses and the responses fetched by brute force or annoy index search.

\section{Conclusion and Future Directions}
We summarize our findings with an example (Table \ref{tab:q2}) to depict the current state of dialogue evaluators and highlight the scope and need for improvement. From our study, we identify the following requirements towards the goal of an ideal evaluator:
\begin{itemize}
\item handle scoring diverse valid responses high
\item sensitivity to grammar and relevance of the response
\item not be heavily influenced by the reference response
\item robust against fooling attacks
\end{itemize}
 In this work we develop a systematic approach to attack a dialogue evaluation system. The attacks point to multiple simple modifications to responses which will have very low correlation with human responses, but still garner high scores from ADEM. Our experiments can be used by the research community to guide a system similar to ADEM towards better performance thereby leading to higher human correlation with the response scores. We believe, from our analysis of ADEM model, that a non-linear scoring model might be better suited for the task of dialogue evaluation. We sincerely hope that the attacks mentioned in this work become the guiding principles for the design of the dialogue evaluation module of the future.
 
 \section{Acknowledgements}
We thank NVIDIA and Robert Bosch Center for Data Sciences and Artificial Intelligence (RBC-DSAI), IIT Madras for compute resources. We also thank Madhuri, Govind, Samprit, Nikhil and Baladitya for helping us with human evaluations.
\bibliography{Bibliography-File}

\begin{thebibliography}{}

\bibitem[\protect\citeauthoryear{Banerjee and
  Lavie}{2005}]{DBLP:conf/acl/BanerjeeL05}
Banerjee, S., and Lavie, A.
\newblock 2005.
\newblock {METEOR:} an automatic metric for {MT} evaluation with improved
  correlation with human judgments.
\newblock In {\em IEEvaluation@ACL},  65--72.
\newblock Association for Computational Linguistics.

\bibitem[\protect\citeauthoryear{Callison{-}Burch, Osborne, and
  Koehn}{2006}]{DBLP:conf/eacl/Callison-BurchOK06}
Callison{-}Burch, C.; Osborne, M.; and Koehn, P.
\newblock 2006.
\newblock Re-evaluation the role of bleu in machine translation research.
\newblock In {\em {EACL}}.
\newblock The Association for Computer Linguistics.

\bibitem[\protect\citeauthoryear{Callison{-}Burch}{2009}]{DBLP:conf/emnlp/Callison-Burch09}
Callison{-}Burch, C.
\newblock 2009.
\newblock Fast, cheap, and creative: Evaluating translation quality using
  amazon's mechanical turk.
\newblock In {\em {EMNLP}},  286--295.
\newblock {ACL}.

\bibitem[\protect\citeauthoryear{Chandrahas, Sharma, and
  Talukdar}{2018}]{Chandrahas2018}
Chandrahas; Sharma, A.; and Talukdar, P.
\newblock 2018.
\newblock Towards understanding the geometry of knowledge graph embeddings.
\newblock In {\em {ACL}}.
\newblock {ACL}.

\bibitem[\protect\citeauthoryear{Galley \bgroup et al\mbox.\egroup
  }{2015}]{DBLP:conf/acl/GalleyBSJAQMGD15}
Galley, M.; Brockett, C.; Sordoni, A.; Ji, Y.; Auli, M.; Quirk, C.; Mitchell,
  M.; Gao, J.; and Dolan, B.
\newblock 2015.
\newblock deltableu: {A} discriminative metric for generation tasks with
  intrinsically diverse targets.
\newblock In {\em {ACL} {(2)}},  445--450.
\newblock The Association for Computer Linguistics.

\bibitem[\protect\citeauthoryear{Gao, Galley, and
  Li}{2018}]{DBLP:journals/corr/abs-1809-08267}
Gao, J.; Galley, M.; and Li, L.
\newblock 2018.
\newblock Neural approaches to conversational {AI}.
\newblock {\em CoRR} abs/1809.08267.

\bibitem[\protect\citeauthoryear{Hori and
  Hori}{2017}]{DSTC6_End-to-End_Conversation_Modeling}
Hori, C., and Hori, T.
\newblock 2017.
\newblock End-to-end conversation modeling track in dstc6.
\newblock {\em arXiv:1706.07440}.

\bibitem[\protect\citeauthoryear{Jia and Liang}{2017}]{DBLP:conf/emnlp/JiaL17}
Jia, R., and Liang, P.
\newblock 2017.
\newblock Adversarial examples for evaluating reading comprehension systems.
\newblock In {\em {EMNLP}},  2021--2031.
\newblock Association for Computational Linguistics.

\bibitem[\protect\citeauthoryear{Li \bgroup et al\mbox.\egroup
  }{2016a}]{N16-1014}
Li, J.; Galley, M.; Brockett, C.; Gao, J.; and Dolan, B.
\newblock 2016a.
\newblock A diversity-promoting objective function for neural conversation
  models.
\newblock In {\em Proceedings of the 2016 Conference of the North American
  Chapter of the Association for Computational Linguistics: Human Language
  Technologies},  110--119.
\newblock Association for Computational Linguistics.

\bibitem[\protect\citeauthoryear{Li \bgroup et al\mbox.\egroup
  }{2016b}]{DBLP:journals/corr/LiZSWZL16}
Li, W.; Zhang, Y.; Sun, Y.; Wang, W.; Zhang, W.; and Lin, X.
\newblock 2016b.
\newblock Approximate nearest neighbor search on high dimensional data -
  experiments, analyses, and improvement (v1.0).
\newblock {\em CoRR} abs/1610.02455.

\bibitem[\protect\citeauthoryear{Li \bgroup et al\mbox.\egroup
  }{2017}]{li2017adversarial}
Li, J.; Monroe, W.; Shi, T.; Ritter, A.; and Jurafsky, D.
\newblock 2017.
\newblock Adversarial learning for neural dialogue generation.
\newblock {\em arXiv preprint arXiv:1701.06547}.

\bibitem[\protect\citeauthoryear{Li, Monroe, and
  Jurafsky}{2016}]{DBLP:journals/corr/LiMJ16a}
Li, J.; Monroe, W.; and Jurafsky, D.
\newblock 2016.
\newblock Understanding neural networks through representation erasure.
\newblock {\em CoRR} abs/1612.08220.

\bibitem[\protect\citeauthoryear{Liang \bgroup et al\mbox.\egroup
  }{2018}]{DBLP:conf/ijcai/0002LSBLS18}
Liang, B.; Li, H.; Su, M.; Bian, P.; Li, X.; and Shi, W.
\newblock 2018.
\newblock Deep text classification can be fooled.
\newblock In {\em {IJCAI}},  4208--4215.
\newblock ijcai.org.

\bibitem[\protect\citeauthoryear{Lin}{2004}]{lin2004rouge}
Lin, C.-Y.
\newblock 2004.
\newblock Rouge: A package for automatic evaluation of summaries.
\newblock {\em Text Summarization Branches Out}.

\bibitem[\protect\citeauthoryear{Liu \bgroup et al\mbox.\egroup
  }{2016}]{DBLP:conf/emnlp/LiuLSNCP16}
Liu, C.; Lowe, R.; Serban, I.; Noseworthy, M.; Charlin, L.; and Pineau, J.
\newblock 2016.
\newblock How {NOT} to evaluate your dialogue system: An empirical study of
  unsupervised evaluation metrics for dialogue response generation.
\newblock In {\em {EMNLP}},  2122--2132.
\newblock The Association for Computational Linguistics.

\bibitem[\protect\citeauthoryear{Lowe \bgroup et al\mbox.\egroup
  }{2015}]{DBLP:conf/sigdial/LowePSP15}
Lowe, R.; Pow, N.; Serban, I.; and Pineau, J.
\newblock 2015.
\newblock The ubuntu dialogue corpus: {A} large dataset for research in
  unstructured multi-turn dialogue systems.
\newblock In {\em {SIGDIAL} Conference},  285--294.
\newblock The Association for Computer Linguistics.

\bibitem[\protect\citeauthoryear{Lowe \bgroup et al\mbox.\egroup
  }{2017}]{DBLP:conf/acl/LoweNSABP17}
Lowe, R.; Noseworthy, M.; Serban, I.~V.; Angelard{-}Gontier, N.; Bengio, Y.;
  and Pineau, J.
\newblock 2017.
\newblock Towards an automatic turing test: Learning to evaluate dialogue
  responses.
\newblock In {\em {ACL} {(1)}},  1116--1126.
\newblock Association for Computational Linguistics.

\bibitem[\protect\citeauthoryear{Mikolov \bgroup et al\mbox.\egroup
  }{2013}]{DBLP:conf/nips/MikolovSCCD13}
Mikolov, T.; Sutskever, I.; Chen, K.; Corrado, G.~S.; and Dean, J.
\newblock 2013.
\newblock Distributed representations of words and phrases and their
  compositionality.
\newblock In {\em {NIPS}},  3111--3119.

\bibitem[\protect\citeauthoryear{Nema and
  Khapra}{2018}]{DBLP:journals/corr/abs-1808-10192}
Nema, P., and Khapra, M.~M.
\newblock 2018.
\newblock Towards a better metric for evaluating question generation systems.
\newblock {\em CoRR} abs/1808.10192.

\bibitem[\protect\citeauthoryear{Papineni \bgroup et al\mbox.\egroup
  }{2002}]{DBLP:conf/acl/PapineniRWZ02}
Papineni, K.; Roukos, S.; Ward, T.; and Zhu, W.
\newblock 2002.
\newblock Bleu: a method for automatic evaluation of machine translation.
\newblock In {\em {ACL}},  311--318.
\newblock {ACL}.

\bibitem[\protect\citeauthoryear{Pennington, Socher, and
  Manning}{2014}]{DBLP:conf/emnlp/PenningtonSM14}
Pennington, J.; Socher, R.; and Manning, C.~D.
\newblock 2014.
\newblock Glove: Global vectors for word representation.
\newblock In {\em {EMNLP}},  1532--1543.
\newblock {ACL}.

\bibitem[\protect\citeauthoryear{Rajpurkar \bgroup et al\mbox.\egroup
  }{2016}]{DBLP:conf/emnlp/RajpurkarZLL16}
Rajpurkar, P.; Zhang, J.; Lopyrev, K.; and Liang, P.
\newblock 2016.
\newblock Squad: 100, 000+ questions for machine comprehension of text.
\newblock In {\em {EMNLP}},  2383--2392.
\newblock The Association for Computational Linguistics.

\bibitem[\protect\citeauthoryear{Ritter, Cherry, and
  Dolan}{2010}]{DBLP:conf/naacl/RitterCD10}
Ritter, A.; Cherry, C.; and Dolan, B.
\newblock 2010.
\newblock Unsupervised modeling of twitter conversations.
\newblock In {\em {HLT-NAACL}},  172--180.
\newblock The Association for Computational Linguistics.

\bibitem[\protect\citeauthoryear{Ritter, Cherry, and
  Dolan}{2011}]{DBLP:conf/emnlp/RitterCD11}
Ritter, A.; Cherry, C.; and Dolan, W.~B.
\newblock 2011.
\newblock Data-driven response generation in social media.
\newblock In {\em {EMNLP}},  583--593.
\newblock {ACL}.

\bibitem[\protect\citeauthoryear{Saha, Khapra, and
  Sankaranarayanan}{2018}]{DBLP:conf/aaai/SahaKS18}
Saha, A.; Khapra, M.~M.; and Sankaranarayanan, K.
\newblock 2018.
\newblock Towards building large scale multimodal domain-aware conversation
  systems.
\newblock In {\em {AAAI}}.
\newblock {AAAI} Press.

\bibitem[\protect\citeauthoryear{Sordoni \bgroup et al\mbox.\egroup
  }{2015}]{DBLP:conf/naacl/SordoniGABJMNGD15}
Sordoni, A.; Galley, M.; Auli, M.; Brockett, C.; Ji, Y.; Mitchell, M.; Nie, J.;
  Gao, J.; and Dolan, B.
\newblock 2015.
\newblock A neural network approach to context-sensitive generation of
  conversational responses.
\newblock In {\em {HLT-NAACL}},  196--205.
\newblock The Association for Computational Linguistics.

\bibitem[\protect\citeauthoryear{Springenberg \bgroup et al\mbox.\egroup
  }{2014}]{DBLP:journals/corr/SpringenbergDBR14}
Springenberg, J.~T.; Dosovitskiy, A.; Brox, T.; and Riedmiller, M.~A.
\newblock 2014.
\newblock Striving for simplicity: The all convolutional net.
\newblock {\em CoRR} abs/1412.6806.

\bibitem[\protect\citeauthoryear{Vinyals and
  Le}{2015}]{DBLP:journals/corr/VinyalsL15}
Vinyals, O., and Le, Q.~V.
\newblock 2015.
\newblock A neural conversational model.
\newblock {\em CoRR} abs/1506.05869.

\bibitem[\protect\citeauthoryear{Wen \bgroup et al\mbox.\egroup
  }{2015}]{DBLP:conf/emnlp/WenGMSVY15}
Wen, T.; Gasic, M.; Mrksic, N.; Su, P.; Vandyke, D.; and Young, S.~J.
\newblock 2015.
\newblock Semantically conditioned lstm-based natural language generation for
  spoken dialogue systems.
\newblock In {\em {EMNLP}},  1711--1721.
\newblock The Association for Computational Linguistics.

\bibitem[\protect\citeauthoryear{Yu \bgroup et al\mbox.\egroup
  }{2017}]{yu2017seqgan}
Yu, L.; Zhang, W.; Wang, J.; and Yu, Y.
\newblock 2017.
\newblock Seqgan: Sequence generative adversarial nets with policy gradient.
\newblock In {\em AAAI},  2852--2858.

\bibitem[\protect\citeauthoryear{Zhao, Dua, and
  Singh}{2018}]{zhao2018generating}
Zhao, Z.; Dua, D.; and Singh, S.
\newblock 2018.
\newblock Generating natural adversarial examples.
\newblock {\em International Conference on Learning Representations}.

\end{thebibliography}
\bibliographystyle{aaai}
\end{document}